\documentclass[11pt]{article}
\usepackage{coling2020}
\usepackage{times}
\usepackage{url}
\usepackage{latexsym}

\usepackage[hidelinks]{hyperref}
\usepackage[utf8]{inputenc}
\usepackage{graphicx}
\usepackage{amsmath}
\usepackage{amsthm}
\usepackage{booktabs}
\usepackage{algorithm}
\usepackage{algorithmic}
\usepackage{amssymb}
\usepackage{multirow}
\usepackage{xcolor}
\usepackage{comment}
\usepackage{overpic}
\usepackage{color}
\usepackage{enumerate}
\usepackage{makecell}
\usepackage{setspace}

\colingfinalcopy % Uncomment this line for the final submission

\title{Logic-guided Semantic Representation Learning for Zero-Shot Relation Classification}

\author{
Juan Li$^{1,2}$\thanks{\quad Equal contribution and shared co-first authorship.}, 
Ruoxu Wang$^{1,2*}$,
Ningyu Zhang$^{1,2*\dag}$, 
Wen Zhang$^{1,2}$, \\
\textbf{Fan Yang}$^{1,2}$, 
\textbf{Huajun Chen}$^{1,2}$ \thanks{\quad Corresponding author} \\
$^{1}$ Zhejiang University \\
$^{2}$ AZFT Joint Lab for Knowledge Engine \\
 {\tt \{lijuan18,ruoxuwang,zhangningyu,wenzhang2015,21821249,huajunsir\}@zju.edu.cn}\\
}

\date{}

\begin{document}
\maketitle
\begin{abstract}
 Relation classification aims to extract semantic relations between entity pairs from the sentences. However, most existing methods can only identify seen relation classes that occurred during training. To recognize unseen relations at test time, we explore the problem of zero-shot relation classification. Previous work regards the problem as reading comprehension or textual entailment, which have to rely on artificial descriptive information to improve the understandability of relation types. Thus, rich semantic knowledge of the relation labels is ignored. In this paper, we propose a novel logic-guided semantic representation learning model for zero-shot relation classification. Our approach builds connections between seen and unseen relations via implicit and explicit semantic representations with knowledge graph embeddings and logic rules. Extensive experimental results demonstrate that our method can generalize to unseen relation types and achieve promising improvements. 
%  Code and dataset are available in \url{anonymous}
\end{abstract}

\section{Introduction}
Relation Classification (RC) is an important task in information extraction, aiming to extract the relation between two given entities based on their related context. RC has attracted increasing attention due to its broad applications in many downstream tasks, such as knowledge base construction \cite{DBLP:conf/emnlp/LuanHOH18} and question answering \cite{DBLP:conf/acl/YuYHSXZ17}.

\begin{figure}
    \vspace{-2mm}
    \centering
    \includegraphics[width=0.8\linewidth, height=200pt]{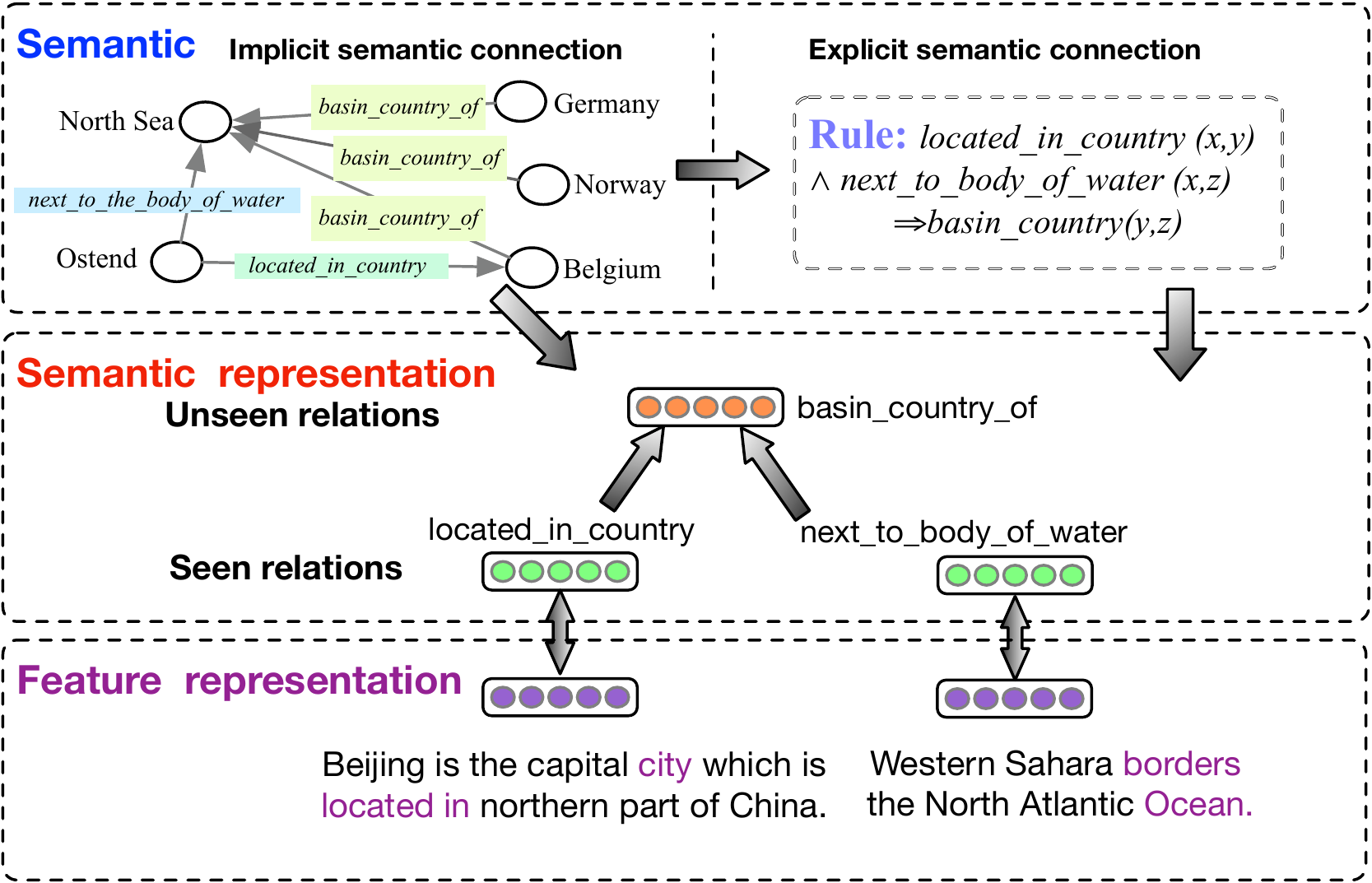}
    \caption{Knowledge graph embedding and rule learning for zero-shot relation classification.}
    \label{instance}
    % \vspace{-2mm}
    \vspace{-5mm}
\end{figure}

Conventional supervised RC approaches can not satisfy the practical needs of the relation classification. In the real world, there exist massive amounts of fine-grained relations. And, the labeled relation types are limited, and each type usually has a certain number of labeled samples.  Naturally, it is prohibitive to generalize to new (unseen) relations (i.e., the model will fail when predicting a type with no training examples). For example, in Figure \ref{instance}, \emph{basin\_country} is an unseen relation type with no labeled sentence in the training stage. To this end, it is urgent for models to be able to extract relations in a zero-shot scenario. 

Previous zero-shot relation classification (ZSRC) approaches leverage transfer learning procedures by reading comprehension \cite{levy2017zero4reading}, textual entailment \cite{obamuyide2018zero4ential}, and so on. However, those methods have to rely on artificial descriptive information to improve the understandability of relation types. Inspired by the zero-shot learning in computer vision \cite{Palatucci2009soc}, it is natural to learn a mapping from the feature space of input samples to the semantic space such as class labels through a projection function. The hypothesis is to build the \textbf{semantic connections} between seen and unseen relations. Conventional approaches usually leverage word embeddings \cite{word2vec} of labels as a common semantic space. We argue that for relation classification, rich semantic knowledge is neglected in the relation labels space: 

\textbf{Implicit Semantic Connection with Knowledge Graph Embedding.} Previous studies \cite{distmult} have shown that the Knowledge Graph Embeddings (KGEs) of semantically similar relations are located near each other in the latent space. For instance, the relation \emph{place\_lived} and \emph{nationality} are more relevant, whereas the relation \emph{profession} has less correlation with the former two relations. Thus, it is natural to leverage this knowledge from KGs to build connections between seen and unseen relations. 

\textbf{Explicit Semantic Connection with Rule Learning.} We human can easily recognize unseen relations via symbolic reasoning. As the example shown in Figure \ref{instance}, with the rule that \textit{basin\_country\_of(y,z)} can be deduced if \textit{located\_in\_country(x,y)} and \textit{next\_to\_body\_of\_water(x,z)}, we can recognize the unseen relation \textit{basin\_country\_of} based on seen relations \textit{located\_in\_country} and \textit{next\_to\_body\_of\_water}. To this end, it is intuitive to infuse rule knowledge to bridge the connections between 
seen and zero-shot relations.

Motivated by this, we take the first step to propose a novel approach, namely, \textbf{L}ogic-guided \textbf{S}emantic \textbf{R}epresentation \textbf{L}earning (\textbf{LSRL}) for zero-shot relation classification. To begin with, we propose to utilize pre-trained knowledge graph embedding such as TransE \cite{transe} to build the \emph{implicit} semantic connection. KGE embeds entities and relations into a continuous semantic vector space and can capture semantic connections between relations in semantic space. Further, we leverage logic rules mined via AMIE \cite{galarraga2013amie} from the knowledge graph and introduce rule-guided representation learning to obtain \emph{explicit} semantic connection. It should be noted that our approach is model-agnostic, and therefore orthogonal to existing approaches. We integrate our approach with two well-known zero-shot methods, namely DeViSE \cite{frome2013devise} and ConSE \cite{norouzi2013conse}. Extensive experimental results demonstrate the efficacy of our approach. 

The main contributions of this work are as follows:
\begin{itemize}
\vspace{-1.5mm}
\item We introduce implicit semantic connection with knowledge graph embedding and explicit semantic connection with rule learning for zero-shot relation classification. 

\item We propose a novel rule-guided semantic representation learning to build connections between the seen and unseen relations. Our work is model-agnostic and can be plugged into different kinds of zero-shot learning approaches.

\vspace{-1.5mm}
\item Extensive experimental results show the efficacy of our approach and also reveals the usefulness of knowledge graph embedding and rule learning. 
\vspace{-1mm}
\end{itemize}

\section{Related Work}

\textbf{Relation Classification.} 
Relation classification (RC) has been firstly proposed in MUC 1998, which aims to predict the relation between two entities by a specific context. Many mature models have been developed to figure out this problem, including traditional methods like \cite{guodong2005exploring}, deep neural networks approach like \cite{zeng2015pcnn,zhang2018attention,zhang2019long,deng2020meta,deng2020low,zhang2020relation,zhang2020can,yu2020devil,wang2020finding}, and some joint models like \cite{zheng2017joint,ye2020contrastive}. However, those methods are all supervised approaches which can only infer relations existing in the train set but are incapable of making predictions for newly-add relations. 

Zero-shot relation classification (ZSRC) was first proposed by \cite{levy2017zero4reading}, which is able to extract new relations by reducing relation classification to answering simple reading comprehension questions. Lately, \cite{obamuyide2018zero4ential} formulates relation extraction as a textual entailment problem and considers the input instance and relation description as the premise and hypothesis. However, these approaches require human annotators to construct questions or write descriptions for relations, which is labor-intensive. On the contrary, our zero-shot relation classification approach does not need any human involvement and can be integrated into most existing RC models.

\textbf{Zero-shot Learning} 
In the computer vision field, zero-shot learning (ZSL) has attracted a lot of attention. The key that underpins ZSL in image recognition is to exploit the shared semantic representations between seen and unseen classes and transfer them to the visual representations of samples.
\cite{frome2013devise} proposes a ZSL model called DeViSE to learn a linear mapping between image features and semantic space using an efficient ranking loss formulation.
\cite{norouzi2013conse} proposes ConSE, which first predicts seen class posteriors, then projects image features into the word2vec space by considering the convex combination of top $T$ most possible seen classes.
The semantic representation of those approaches is learned by certain auxiliary information attached to the class labels, such as attribute description \cite{Jayaraman2014attribute2014,Farhadi2009attribute1} and embedding representation \cite{romera2015embarrassingly,Akata2016ale}.
Different from the zero-shot approaches in computer vision, we construct the semantic space by considering information from the knowledge graph rather than word embedding or attribute.  

There are also some ZSL applications in NLP, such as event extraction \cite{huang2017zero4ee}, entity-typing \cite{zhou2018zero4entitytyping}, cross-lingual entity linking \cite{Rijhwani2018entitylinking}, text classification \cite{Pushp2017textclassification} and cold-start recommendation \cite{zero4rc}, as well as in KG such as link prediction\cite{zsgan20}.

\textbf{Knowledge Graph Embedding.}
In recent years, various KG embedding methods, including translation-based, semantic matching and neural network methods, have been devised to learn vector representations for entities and relations of a KG. 
Translation-based models \cite{transe,transd,transr} use distance-based scoring functions to assess the plausibility of a triple. For example, in TransE \cite{transe}, the score function is $f_r(h,t)=||h+r-t||^2_{l_{1/2}}$. Semantic matching models \cite{distmult,hole,analogy} employ similarity-based scoring functions to compute the energy of relational triples, where the scoring function of the representative model DistMult \cite{distmult} is $f_r(h,t)=h^{\top}diag(M_r)t$. Neural network models learn to express entities and relations through neural networks, such as CNN-based methods \cite{conve} and GNN-based methods \cite{r-gcn}. 
We utilize KG embedding models to learn the representations of relations instead of word embeddings so that the representations of relations are only related to the structure of a KG but not relations' name. Meanwhile, connections of relations are harvested. 
% Considering TransE, under the assumption that $h+r=t$, connections of relations are harvested. We can get $r_1 + r_2 = r_3$ if rule $r_1(x, y) \land r_2(y, z) \Rightarrow r_3(x, z)$ exists, as mentioned in \cite{lin2015pathemb}.

\textbf{Rule Learning.}
Rules over a KG can capture connections between relations, and a variety of methods for rule learning have been studied, such as Inductive Logic Programming(ILP) algorithms, rule mining methods, and embedding-based methods.  
ILP is formalized by first-order logic and has strong representation powers, but does not scale to large datasets \cite{drum2019nips}. To address this, several efficient rule miners for KGs have been developed, such as RDF2rules \cite{rdf2rules}, ScaleKB \cite{scalekb} and AMIE+ \cite{amie+}. 
In addition, embedding-based rule learning methods have gained attention. RLvLR \cite{rlvlr} utilizes embeddings to guide rule extraction and reduce the search space. DistMult \cite{distmult} utilizes learned embeddings of entities and relations to extract logical rules. And \cite{DBLP:conf/semweb/HoSGKW18} introduces a framework for rule learning guided by external sources.   
We adopt the widely used rule mining method proposed in \cite{galarraga2013amie} to extract rules from KG in this paper.
To the best of our knowledge, we are the first approach to address zero-shot relation classification with the assistance of logical rules from KG. 

\section{Methodology}
\subsection{Preliminaries}
We start by defining some notations and terms. $\mathcal{R}^S$ denotes the set of seen relations during training and $\mathcal{R}^U$ denotes the set of unseen relations for testing, and $\mathcal{R}^S \bigcap \mathcal{R}^U=\phi$. 
$\mathcal{D}^{tr}={\{(s_i, h_i, t_i, y_i),i=1,...,N_{s}\}}$ denotes training dataset, where $s_i$ represents one sentence and $(h_i,t_i)$ is an entity pair mentioned in $s_i$. $N_{s}$ is the number of seen relations and $y_i \in \mathcal{R}^S$ denotes the relation of the entity pair. Analogously, $\mathcal{D}^{ts}= {\{(s_j, h_j, t_j, y_j),j=1,...,N_{u}\}}$ denotes the test dataset, where $N_{u}$ is the number of unseen relation and $y_j \in \mathcal{R}^U$ denotes the relation of $h_j$ and $t_j$ in the sentence $s_j$. 

\begin{figure}[htbp]
    \centering
    \includegraphics[width=0.8\linewidth, height=240pt]{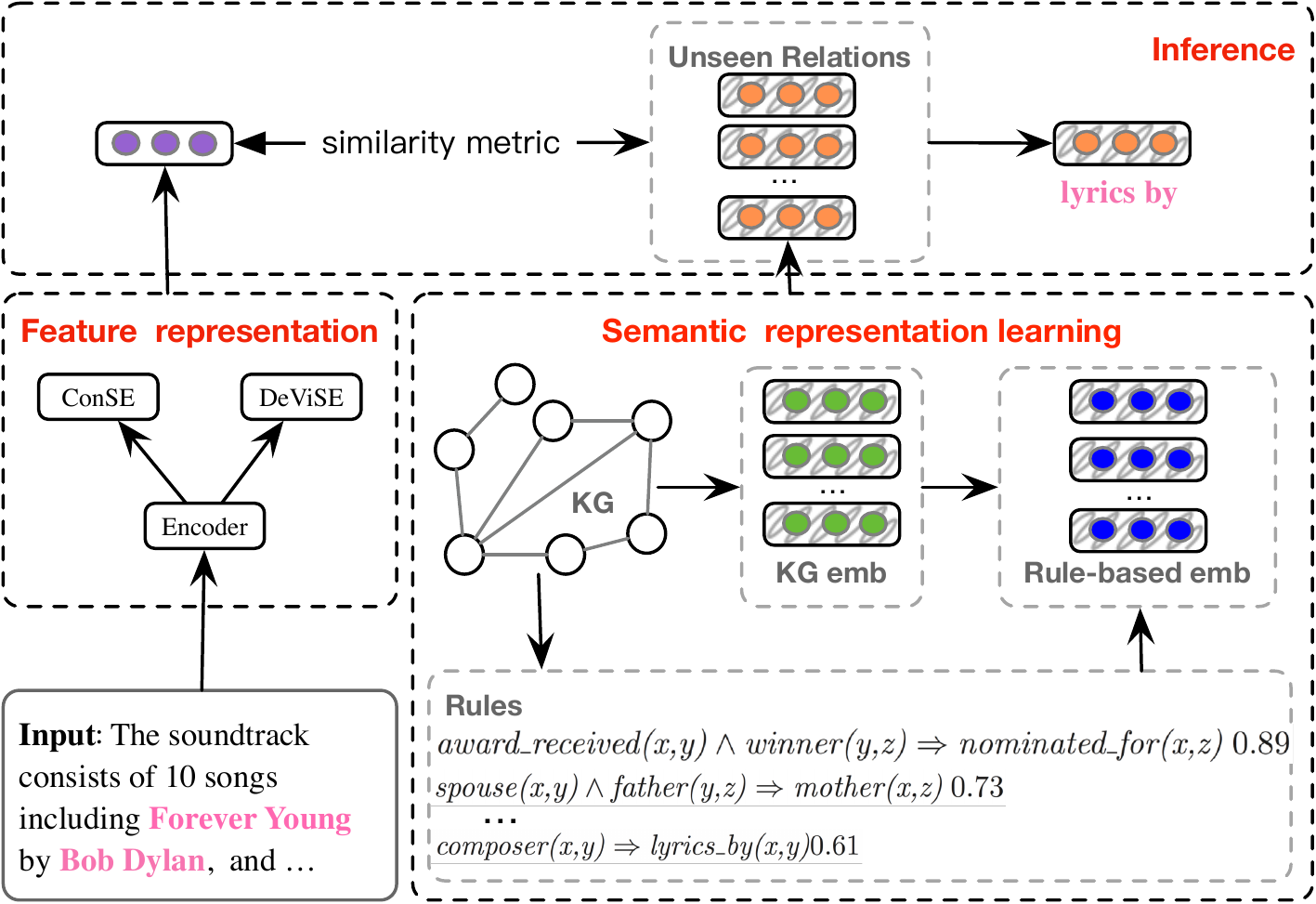}
    \caption{The architecture of Logic-guided Semantic Representation Learning model.} 
    \label{zsrl-model}
\end{figure}

The overall framework is illustrated in Figure \ref{zsrl-model}, which is composed of three modules: 

\textbf{Feature Representation} (\S \ref{encoder_sec}) encodes the input sentence into the feature space, which is aimed at capturing syntax features of sentences.

\textbf{Semantic Representation Learning} (\S \ref{semantic_sec}) maps relation types into a semantic space and builds up the connections between seen and unseen relations. Specifically, we propose logic-guided semantic representation learning with knowledge graph embedding and rule learning.

\textbf{Inference} (\S \ref{infer_sec}) predicts relation types via computing similarity between feature representation of current input sentence and semantic representations for all unseen relations. We infer the unseen relation with the label that is most similar in the semantic space.

\subsection{Feature Representation\label{encoder_sec}}
The input of feature representation is a sentence, and the output is its vector representation. Firstly, we use the Piecewise Convolutional Neural Networks(PCNNs) \cite{zeng2015pcnn} model to encode input instance, and then use two types of projection functions including \textit{DeViSE} and \textit{ConSE} to get the final feature representation of the input instance.

PCNNs has been proven to be effective in RC. It inputs the concatenation of word embedding and position embedding into a Convolution Neural Network(CNN) to obtain the hidden layer representation $h$. Then, $h$ is divided into three parts based on the two entities' positions, and max pooling is perfomed on each part to obtain $(pl_1,pl_2,pl_3)$. Final feature encoding of the input sentence $f = [pl_1;pl_2;pl_3]$ is the concatenation of the three pooling segments. We denote the process as below:
\begin{equation}
   f = PCNN(x_1,...,x_n)
   \vspace{-1mm}
\end{equation}
\textit{DeViSE} formulates ZSL as a regression problem and learns a linear function to project the input representation to target semantic space:
\begin{equation}
    g = W *f+b
    \label{devise formulation}
\end{equation}
\textit{ConSE} maps input representation into target semantic space via convex combination. It trains a classifier $C$ on training dataset $D^{tr}$ and obtains top $T$ probable seen relation types $R_t^S$ together with their probability $p_t$. $E(R_t^S)$ is the embedding of $R_t^S$, and then weighted sum on $E(R_t^S)$ is regarded as the feature representation of inputs. The process can be formulated as follows:
\vspace{-1mm}
\begin{equation}
    \label{top_t equation}
    R^S_t, p_t, E(R^S_t)= C(f), t=1,...,T
    \vspace{-5mm}
\end{equation}

\begin{equation}
    \label{weight-sum equation}
    g = \sum _{t=1}^T p_t*E(R_t^S)
    \vspace{-1mm}
\end{equation}

\subsection{Semantic Representation Learning\label{semantic_sec}}
Semantic representation builds connections between unseen and seen relations in ZSRC via external resources. We describe the following three kinds of embedding representations in a semantic space.

\textbf{Word Embedding} denoted as $E_{wd}$ is the commonly used method. However, this way faces challenges as analyzed in the introduction. In order to capture the rich explicit or implicit semantic connection between relations, two forms of embedding methods based on KG are introduced. 

\textbf{KG Embedding} embeds relations and entities into latent low-dimensional continuous-space vectors, denoted as $E_{kg}$. 
In KG embedding methods, the score of a triple (h,r,t) can be calculated via head entity embedding E(h), relation embedding E(r), and tail embedding E(t). Different methods follow different assumptions. For example, the typical method TransE \cite{transe} supposes a translation law in the semantic space where $E(h) + E(r) = E(t)$ for positive triples in a KG. Therefore, relation embedding from KG embedding methods is related to triples it belongs to and is not affected by what words it contains. 
While sometimes the word contained in the relation string also reveals the semantic of this relation, word embedding and KG embedding can complement each other at that time. Hence, we also consider combining them together through a linear transformation, \textbf{KG+Word embedding} is defined as:
\begin{equation}
  E_{kw} = W_2 * ([E_{kg}; E_{wd}] + b_2)
  \label{kg+word formula}
\end{equation}
where $[x;y]$ means concatenation of $x$ and $y$.

\textbf{Rule-guided Embedding} represents rules in vector space instead of symbolization, denoted as $E_{rl}$. 
In a knowledge graph, logic rules show the connections between relations. They are in the form of \textit{body $\Rightarrow$ head}, where \textit{head} is a binary atom and \textit{body} is a conjunction of binary and unary atoms, such as rule \textit{spouse(x,y) $\wedge$ father(y,z) $\Rightarrow$ mother(x,z)}, and the number of atoms in \textit{body} is the length of the corresponding rule. 
We adopt typical rule mining methods such as AMIE \cite{galarraga2013amie} to generate rules from structural KGs. In addition to rules, AMIE also produces the PCA confidence $conf$ to filter out rules. Inspired by \cite{zhang2019iteratively}, we apply an simple but effective embedding-based method to incorporate symbolic rules into semantic space and generate $E_{rl}$. Taking TransE as an example, it assumes  $h +r \approx t$ for a positive triple $(h,r,t)$. According to this assumption, we can get $r_1 + r_2 = r_3$ if rule $r_1(x, y) \land r_2(y, z) \Rightarrow r_3(x, z)$ exists as mentioned in \cite{lin2015pathemb}. Thus if the rule contains an unseen relation, embedding of the unseen relation can be calculated based on other seen relations in this rule. Besides, it is possible that one unseen relation involves multiple rules, for that, we calculate unseen relation's embedding as follows: 
\begin{equation}
    E_{rl}(R_i^U) = \frac { \sum _{j=1}^K conf_j*E_{kg}(Rule_{ij}^U) } {\sum _{j=1}^K conf_j}
    \vspace{-1mm}
\end{equation}
where $R_i^U$ represents the $i_{th}$ unseen relation, $Rule_{ij}^U$ is the $j_{th}$ rule in the set of rules about $R_i^U$ with top $K$ highest PCA confidence score and $conf_j$ represents the PCA confidence of $Rule_{ij}^U$.
For example, with two rules about unseen relation $r$, $R1: r_A \wedge r_B \Rightarrow r$ and $R2: r_C \wedge r \Rightarrow r_D$, following TransE's assumption, we calculate embedding of $r$ via $E_{rl}(r) = \frac{conf_1 * [E_{kg}(r_A) + E_{kg}(r_B)]+conf_2 * [E_{kg}(r_D) - E_{kg}(r_C)]}{conf_1 + conf_2}$.

Similar to Word+KG embedding, we also consider \textbf{KG+Rule Embedding}, denoted as $E_{kr}$, as they might also complement each other. $E_{kr}$ can be calculated as follows:
\vspace{-1mm}
\begin{equation}
    E_{kr} = \lambda * E_{rl} + (1-\lambda)*E_{kg}
    \label{kg+rule formula}
    \vspace{-1mm}
\end{equation}
where $\lambda$ is a hyperparameter representing the combination weight between KG embedding and rules. It is set as 0.5 in our experiment. 
Meanwhile, we calculate \textbf{Rule+Word Embedding}, denoted as $E_{rw}$, by replacing $E_{kg}$ in Equation \ref{kg+word formula} with  $E_{rl}$.

\subsection{Inference\label{infer_sec}}
During prediction, we compare the similarity between feature representation $f$ of input sentence and the semantic representation of unseen relations as follows:
\begin{equation}
    \label{similarity equation}
     \overline{y_i}=\mathop{sim}(f_{x_i},E(R^U_{x_i}))
     \vspace{-1mm}
\end{equation}
The similarity function $sim( )$ can be cosine similarity or Euclidean Distance. 
% During training, the output similarity $\overline{y_i}$ of each sample $x_i$ in $D^{tr}$ should close to its relation type $y_i$. Training process varies under different projection functions. For DeViSE, we obtain the type with the highest scores by calculating the similarity between feature representation and relation embedding. 
% We then adopt cross-entropy with $l2$ regularization as loss function to train the model \cite{zeng2015pcnn}. For ConSE, the training object is to minimize cross-entropy loss of the classifier $C$ in Equation \ref{top_t equation}. 
Unseen relations with higher similarity with sentence feature representation are more likely to be predicted.

\section{Experiment}
In experiments, we want to explore: 1) Whether embeddings based on KGs are more useful for the ZSRC task than word embeddings? 2) What is the factor that can strengthen semantic representations in ZSRC? 3) Whether and how can logical knowledge help build better semantic space?

\subsection{Datasets}
\label{dataset}
Different from zero-shot learning relation classification dataset of \cite{levy2017zero4reading} and \cite{obamuyide2018zero4ential}, our method considers rules of relations during training rather than question templates or relation descriptions. We construct a new dataset based upon Wikipedia-Wikidata \cite{sorokin2017context} relation extraction dataset which contains 353 relations and 856,217 instances. 
To evaluate the capability of injecting rule logic into the zero-shot prediction models, we ensure that relations have certain connections in our dataset.
We firstly cluster all 353 relations based on word embeddings, then divide seen and unseen relations according to the instance number of relations by the given threshold (1200) for one cluster. 
We drop relations from the cluster where all relations' instance number is less than 500 with the assumption that there is no support from related seen labels. 
Manual adjustments are further applied to get the final dataset. 
In this dataset, there are ultimately 100 relations, 70 of which are seen relations and the rest 30 are unseen ones.
% \zny{dataset table}
% \input{data_wiki.tex}

\subsection{Settings}
\label{experimental settings}
\begin{table}[htb]
\title{Parameter Settings.}
\centering
% \small
\begin{tabular}{p{35pt}|p{120pt}|l|p{35pt}}
\Xhline{1.2pt}
Modules&Parameters&Settings(Tried)&Settings\\
\hline
\multirow{6}*{PCNN}&Kernel Size&{3,4,5}&3\\
% \cline{2-4}
% &Word Embedding Size&100&100\\
\cline{2-4}
&Position Embedding Size&{5,10,15}&5\\
\cline{2-4}
&Learning Rate& {0.01,0.05,0.005} & 0.01\\
\cline{2-4}
&Dropout& {0,0.1,0.3,0.5} &0.5\\
\cline{2-4}
&Channel Number& {150,200,250,300} & 250 \\
\hline
% \Xhline{0.8pt}

% {BERT}&Learing Rate& {0.005,0.001,0.0005} & 0.0005\\
% \hline
{ConSE}&T& {1,2,3} & 3\\
% \Xhline{0.8pt}
\hline
{DeViSE}&Margin& {0.8,1.0,1.2} & 1.0\\
\hline
% {REPT-KGE}&$\lambda$& {0.2,0.5,0.8} & 0.2\\
% \Xhline{0.8pt}
% {REPT-Rule}&$\lambda$& {0.8,1.0,1.2} & 1.0\\
\Xhline{1.2pt}
\end{tabular}
\caption{Parameter Settings}
\label{parameter}
% \vspace{-5mm}
\end{table}

% \textbf{Word Embedding.} 
We use Wikipedia documents\footnote{https://dumps.wikimedia.org/enwiki/latest/enwiki-latest-pages-articles.xml.bz2} to train \textbf{word embeddings}, where words appear more than ten times are preserved in vocabulary, following a common setting as other works. Word2vec \cite{word2vec} is applied for word embedding training with window size set as 5. 
% \textbf{KG Embedding.}
For \textbf{KG embeddings}, we use TransE to train the embedding of entities and relations on Wikidata, which contains about 20,982,733 entities and 594 relations in total. The embedding size is set as 100, the margin as 1.0, and the learning rate as 0.01. 
% \textbf{Other Parameter Settings.} 
For the PCNN layer, we set kernel size as 3, position embedding size as 5, the number of channels as 250, margin as 2.0, learning rate as 0.01, and dropout as 0.5. 
For ConSE, Top $3$ seen classes are chosen for prediction. 
For DeViSE, the margin is set as $1.0$. 
For rule mining, we set the max length of rules as $2$.

\subsection{Whether KG-based embeddings are more useful than word embeddings?}
\begin{table}[!htbp]
\renewcommand\arraystretch{1.1}
	\centering 
	\begin{tabular}{p{1.3cm}<{\centering}|p{1.3cm}<{\centering}|p{1.3cm}<{\centering}|p{1.3cm}<{\centering}|p{1.3cm}<{\centering}|p{1.3cm}<{\centering}|p{1.3cm}<{\centering}}
	  \hline
	  \Xhline{1.0pt}
	  \multirow{2}{*}{} 
				 & \multicolumn{3}{c|}{ConSE(Hit@n)}& \multicolumn{3}{c}{DeViSE(Hit@n)}  \\ \cline{2-7}
				 & 1 & 2 & 5 & 1 & 2 & 5 \\ 
	  \Xhline{1.0pt}
	+$E_{wd}$	& 0.21 & 0.30 & 0.43 & 0.11 & 0.19 & 0.39 \\ 
    \Xhline{0.6pt}
	+$E_{kg}$	& 0.39 & 0.53 & 0.69 & 0.22 & 0.38 & 0.57 \\ 
	+$E_{rl}$	& 0.40 & 0.54 & 0.72 & 0.23 & 0.39 & 0.58 \\ 
	\Xhline{0.2pt}
	+$E_{kw}$	   & 0.39 & 0.55 & 0.72 & 0.23 & \textbf{0.40} & \textbf{0.59} \\ 
    +$E_{rw}$ & 0.40 & 0.55 & 0.70 & 0.23 & 0.34 & 0.57 \\ 
	+$E_{kr}$	   & \textbf{0.43} & \textbf{0.57} & \textbf{0.74} & \textbf{0.25} & 0.39 & \textbf{0.59}  \\
	\Xhline{1.0pt}
 
	\end{tabular}
  \caption{Performance of DeViSE and ConSE in the case of different embedding methods, including Word($E_{wd}$), KG($E_{kg}$), Rule($E_{rl}$), KG+Word($E_{kw}$), Rule+Word($E_{rw}$) and KG+Rule($E_{kr}$) embeddings.
  % 解释图
  }
	\label{overallresults}
\end{table}
To compare the effectiveness of KG-based embeddings and word embeddings, we regard methods that use $E_{wd}$ as baselines. Results of two kinds of methods, including KG-based embeddings and a combination of any two embeddings, are listed to explore the usefulness of KGs in the ZSRC task. The experiments also distinguish the results when using two different projection functions in the ZSL problem, i.e., ConSE and DeViSE.
During testing, we rank the similarity scores between feature representations of test sentences and all unseen relations' semantic representations based on cosine similarity and get ranks of true labels. Hit@K(K=1, 2, 5) are used as evaluation metrics. 
\begin{table*}[!hbt]
\renewcommand\arraystretch{1.1}
	\centering
	\small
    % \normalsize
	\begin{tabular} {p{90pt}|c|c|p{102pt}|p{100pt}}

		\Xhline{1.2pt}

		\multirow{2}{*}{\shortstack{Unseen Relations}}& \multicolumn{2}{|c|}{F1-score} & \multicolumn{2}{|c}{Top $3$ Related Seen Relations}\\ \cline{2-5}
		& \multicolumn{1}{|c|}{+$E_{kg}$} & \multicolumn{1}{|c|}{ +$E_{wd}$} & \multicolumn{1}{|c|}{ +$E_{kg}$ } & \multicolumn{1}{|c}{+$E_{wd}$} \\	
		\Xhline{0.8pt}

		\multirow{3}{*}{lyrics\_by} & \multirow{3}{*}{\textbf{0.52}}& \multirow{3}{*}{0.06}   & performer &influenced\_by \\ \cline{4-5}
		&&& composer &spouse \\ \cline{4-5}
		&&& cast\_member &cast\_member \\
		\Xhline{0.8pt}
		
		\multirow{3}{*}{after\_a\_work\_by}  & \multirow{3}{*}{\textbf{0.51}}& \multirow{3}{*}{0.01}  & author & named\_after \\ \cline{4-5}
		&&& screenwriter & author \\ \cline{4-5}
		&&& creator & characters \\ 
		\Xhline{0.8pt}

		\multirow{3}{*}{\shortstack{location\_of\_formation}}& \multirow{3}{*}{\textbf{0.46}}& \multirow{3}{*}{0.02} & headquarters\_location & subclass\_of \\ \cline{4-5}
		&&& location & opposite\_of\\ \cline{4-5}
		&&& capital & part\_of \\ 
		\Xhline{0.8pt}

		\multirow{3}{*}{nominated\_for} & \multirow{3}{*}{\textbf{0.97}}  & \multirow{3}{*}{0.56}  & award\_received&award\_received  \\ \cline{4-5}
		&&&winner   &part\_of   \\ \cline{4-5}
		&&&participant\_of&member\_of   \\ 
		\Xhline{0.8pt}
		\multirow{3}{*}{mother} & \multirow{3}{*}{0.40}  & \multirow{3}{*}{\textbf{0.83}}  &follows &child  \\ \cline{4-5}
		&&& spouse  &spouse   \\ \cline{4-5}
		&&& twinned\_administrative\_body  &father   \\ 
		\Xhline{0.8pt}

		\multirow{3}{*}{developer} & \multirow{3}{*}{0.38}  & \multirow{3}{*}{\textbf{0.49}}  & publisher&manufacturer  \\ \cline{4-5}
		&&& manufacturer  &publisher   \\ \cline{4-5}
		&&& owned\_by &owned\_by   \\ 
		\Xhline{0.8pt}

		\multirow{3}{*}{office\_contested} & \multirow{3}{*}{\textbf{0.26}}  & \multirow{3}{*}{0.00}  &position\_held &  \\ \cline{4-4}
		&&&successful\_candidate& \quad\quad\quad\quad--------------  \\ \cline{4-4}
		&&&applies\_to\_jurisdiction   &  \\ 
		\Xhline{0.8pt}

		\multirow{3}{*}{occupant}  & \multirow{3}{*}{\textbf{0.31}} & \multirow{3}{*}{0.00}  &owned\_by& \multirow{3}{*}{\quad\quad\quad\quad--------------} \\ \cline{4-4}
		&&&location  &  \\ \cline{4-4}
		&&&headquarters\_location&  \\ 
		\Xhline{0.8pt}

		\multirow{3}{*}{drafted\_by}  & \multirow{3}{*}{\textbf{0.81}} & \multirow{3}{*}{0.00}  & member\_of\_sports\_team & \multirow{3}{*}{\quad\quad\quad\quad--------------}  \\ \cline{4-4}
		&&& educated\_at & \\ \cline{4-4}
		&&& member\_of & \\ 
		% \Xhline{0.8pt}

		% \multirow{3}{*}{} & \multirow{3}{*}{0.}  & \multirow{3}{*}{0.}  & &  \\ \cline{4-5}
		% &&&   &   \\ \cline{4-5}
  %   	&&&   &   \\ \hline

		\Xhline{1.0pt}

	% \multirow{3}{*}{deepest point } & \multirow{3}{*}{0.49} & \multirow{3}{*}{0.19}  & located in or next to body of water &located on terrain feature \\ \cline{4-5}
		% &&&located on terrain feature  & located in or next to body of water  \\ \cline{4-5}
  %   	&&&mouth of the watercourse  & continent \\ 
  %   	\hline
	\end{tabular}
	
	\caption{Results of KG embedding and word embedding on F1 score when using ConSE as projection function. And top 3 most influential seen relations of the corresponding unseen relation are presented.}
	\label{case table}
% 	\vspace{-5mm}
	
\end{table*}

The overall results are shown in Table \ref{overallresults}. 
% From Table \ref{overallresults},
Under the ConSE structure, methods that incorporate semantic representation based on the knowledge graph significantly outperform word embedding. Specifically, KG embedding gains improvement with 18\% and rule embedding with 19\% on Hit@1. 
The performance of word+KG embedding and word+rule-based embedding also improves a lot, and the combination of KG+rule-based embedding achieves the best performance. Thus, we can conclude that \emph{KG-based embeddings are superior to word embedding}. 

Additional inspection of the table shows that the results over ConSE are better than DeViSE in all embedding settings. The reason is associated with the difference between the two models. The representation space of ConSE is limited to a combined space consisting of seen classes, while DeViSE enables mapping instance embedding to the whole relation space. Thus the dataset with stronger relevance between relations is more friendly to ConSE, which is the case 
in our ZSRC dataset.

\subsection{What is the factor that can strengthen semantic representations in ZSRC?}
\label{kg vs word} 
We analyze the question via a comparison between word embeddings and KG embeddings. 
A closer inspection of a specific relation is illustrated in Table \ref{case table}.
For most unseen relations, KG embeddings perform better than word embeddings. Especially for \emph{drafted\_by}, \emph{occupant} and \emph{office\_contested}, word embeddings predict almost nothing, while KG embeddings achieve 81\%, 31\% and 26\% respectively. The reason may be that word embedding is less than enough to capture complete, accurate, or even logic-level connections between relations. For example, for the relation \emph{drafted\_by} which means ``allocate certain players to teams in some sports", it is difficult for word embedding to capture its connection to the relations \emph{member\_of\_sports\_team} and \emph{educated\_at} because the word \emph{draft} has other senses such as ``draft a document" that appears much more commonly in the corpus. By contrast, KG embeddings are trained based upon the entity pairs and relationships existing within a whole knowledge base, thereby capturing the more accurate meaning of, and connections between these relations.

Negative examples are also found in Table \ref{case table}, such as \emph{mother}, for which KG embedding predicts poorly, but word embedding achieves 83\%. The reason may be that the number of training triples for \emph{mother} is relatively small in our dataset, leading to poor embeddings. KG embedding suffers from its sparsity problem due to imperfect KG, whereas word embedding excels at capturing contextually similar words such as \emph{mother} to \emph{father} or \emph{spouse}. 

These results show that \emph{successfully building accurate or even logical-level connections between seen and unseen relations is an essential factor for zero-shot tasks, and this is why KG embeddings perform better than word embeddings.}

\begin{table*}[!htb]
\renewcommand\arraystretch{1.1}
	\centering
	\small
	% \begin{tabular}{p{67pt}|p{25pt}|p{25pt}|p{25pt}|p{280pt}}

	\begin{tabular}{l|p{0.55cm}<{\centering}|p{0.55cm}<{\centering}|p{0.55cm}<{\centering}|p{0.55cm}<{\centering}|p{0.55cm}<{\centering}|p{0.55cm}<{\centering}|l}
		\Xhline{1.2pt}
		  \multirow{2}{*}{\shortstack{Unseen\\Relations}} & \multicolumn{6}{c|}{F1-score} & \multirow{2}{*}{ Related rules w.r.t. unseen relations}\\ \cline{2-7} &+$E_{wd}$ &+$E_{kg}$ &+$E_{rl}$ & +$E_{kw}$ & +$E_{rw}$ &+$E_{kr}$ & \\ 
      \Xhline{0.8pt}

		  \multirow{2}{*}{mother} & \multirow{2}{*}{\textbf{0.83 }} 
      &\multirow{2}{*}{0.40} & \multirow{2}{*}{0.77}
      &\multirow{2}{*}{0.53} 
      &\multirow{2}{*}{0.80} & \multirow{2}{*}{0.78}
			&\textit{mother(x,z)} $\Leftarrow$ \textit{spouse(x,y)} $\wedge$ \textit{father(y,z)}\\ 
			  &&&&&&& \textit{mother(x,y)}   $\Leftarrow$ \textit{child(y,x)}\\ 
      \hline
	  
		  lyrics\_by & 0.06 & 0.52 & 0.51 &
      0.49 & 0.48 & \textbf{0.52} &
      \textit{lyrics\_by(x,y)} $\Leftarrow$ \textit{composer(x,y)} \\
      \hline

		  nominated\_for  & 0.56  & \textbf{0.97} & 0.96
      & \textbf{0.97} & 0.96 & 0.96
			&\textit{nominated\_for(x,z)} $\Leftarrow$ \textit{award\_received(x,y)} $\wedge$ \textit{winner(y,z)}  \\ 
		  \hline

		  % producer & 0.43 & 0.54 & \textbf{0.55}  &   \textcolor[rgb]{0.88,0.31,0.04}{ \textit{screenwriter(x,y)} } $\Rightarrow$ \textit{producer(x,y)}\\ \hline
	  \multirow{3}{*}{producer} & \multirow{3}{*}{0.41}   &
      \multirow{3}{*}{0.52}  & \multirow{3}{*}{\textbf{0.55}}&
      \multirow{3}{*}{0.54}   &\multirow{3}{*}{0.52}  & \multirow{3}{*}{0.53}&
	  \textit{producer(x,y)} $\Leftarrow$ \textit{director(x,y)} \\ 
	   &&&&&&& \textit{producer(x,y)} $\Leftarrow$ \textit{screenwriter(x,y)}\\
	   &&&&&&& \textit{producer(x,y)} $\Leftarrow$ \textit{cast\_member(x,y)}\\
	   \hline
		
		  field\_of\_work&0.04&0.14& 0.29
      & 0.11 & 0.29 & \textbf{0.37}
      & \textit{field\_of\_work(x,y)} $\Leftarrow$ \textit{occupation(x,y)}\\ 
      \hline

		  connecting\_line&0.00& 0.10 & 0.43
      &0.28 & 0.42 & \textbf{0.47}
      & \textit{connecting\_line(x,z)} $\Leftarrow$ \textit{adjacent\_station(y,x)} $\wedge$ \textit{part of(y,z)} \\ 
      \hline

	   \multirow{2}{*}{residence}  & \multirow{2}{*}{0.01} &\multirow{2}{*}{0.32}& \multirow{2}{*}{0.30} &
    \multirow{2}{*}{0.30} &\multirow{2}{*}{0.38}& \multirow{2}{*}{\textbf{0.39}} 
		& \textit{residence(x,y)} $\Leftarrow$ \textit{place\_of\_birth(x,y)}\\ 
		&&&&&&& \textit{residence(x,y)} $\Leftarrow$ \textit{place\_of\_death(x,y)}\\
		  
	  \Xhline{1.2pt} %添加表格底部粗线
	\end{tabular}
	\caption{Results of all different embeddings on F1 score when regrading ConSE as project funtion, and related rules w.r.t unseen relations.
	}
	\label{rule table}
	\vspace{-3mm}
\end{table*}
\subsection{Whether and how can logical rules help build better semantic space?}
\label{rule learning with kg embedding}
We investigate this question via logic rule analysis. 
General inspection of Table \ref{overallresults} reveals that rule-based embedding is slightly better than single KG embedding, and \emph{KG+Rule embedding} achieves the best result with 3$\sim$4\% improvement in overall Hit@1 score under ConSE.
A further examination of case studies over different kinds of semantic representations is listed in Table \ref{rule table}. It shows that most relations based on rule embedding achieve at least comparable results with KG embedding such as \emph{nominated\_for}, \textit{producer}, \textit{lyrics\_by}. Some relations, such as \emph{producer}, outperform KG embedding slightly.  
This may because rule embedding can capture logic-level connections between seen and unseen relations. For example, the unseen relation \emph{nominated\_for} is logically related with two seen relations \emph{award\_received} and \emph{winner} with the rule \textit{nominated\_for(x,z)} $\Leftarrow$ \textit{award\_received(x,y)} $\wedge$ \textit{winner(y,z)}.  
%Other examples such as the seen relation \emph{place\_of\_death} logically entails the unseen relation \emph{residence} to some extend, \emph{screenwriter} somewhat entails \emph{producer}, and so on. 
The most interesting aspect is about the relation \emph{mother} for which KG embedding fails to compare with word embedding because of being poorly trained, while rule embedding achieves comparable scores with word embedding. The reason may be that rule embedding helps strengthen the embedding by incorporating more knowledge from related relations contained in the rules, thus making a correction to the relation embedding.  

\begin{figure}[!htb]
    \centering
    \includegraphics[scale=0.9]{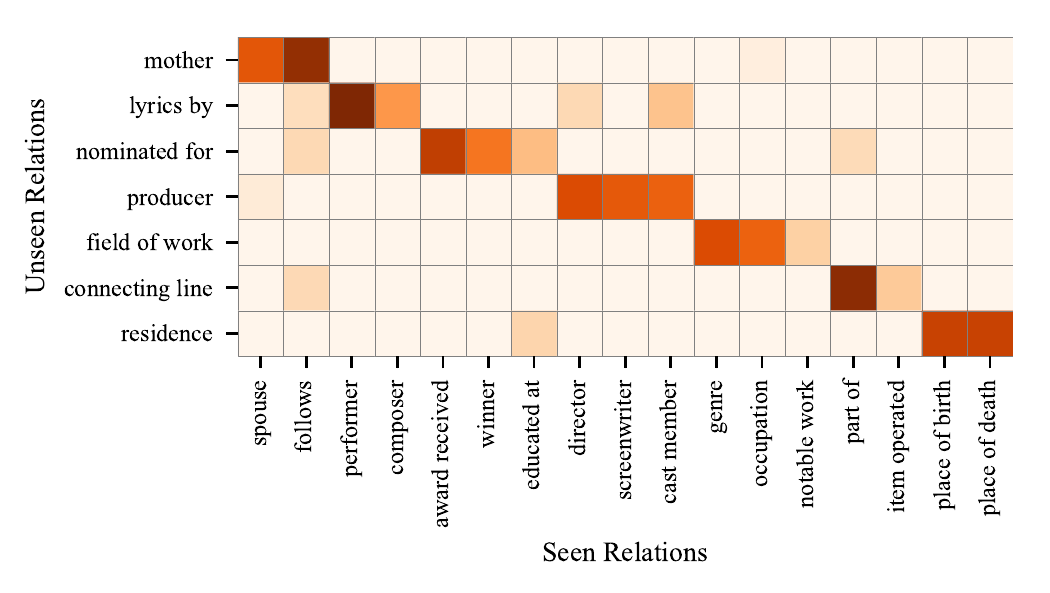}
    \vspace{-4mm}
    \caption{This heatmap is constructed from the result of ConSE+KG, reflecting the incidence of seen relations on unseen relations. Where the horizontal axis represents seen classes and the vertical axis represents unseen classes.
    % Unseen relations are the same as in Table \ref{rule table}.
    }
    \label{heatmap_kg}
    % \vspace{-4mm}
\end{figure}

We also represent a heatmap of ConSE+KG results corresponding to Table \ref{rule table}, as shown in Figure\ref{heatmap_kg}. From the heatmap, we can discover that KG embeddings could capture logical connections for the semantic connections between unseen and seen relations. For example, the relation \emph{award\_received} plays an important role in prediction of the unseen relation \emph{nominated\_for}, which is exactly consistent with the rule { \emph{award\_received(x,y)} $\wedge$ \emph{winner(y,z)} $\Rightarrow$ \emph{nominated\_for(x,z)} } in Table \ref{rule table}. Similar matched correspondence for other relations are 
found. 
These results and analysis show that \emph{logical connections between relations expressed by rules could help build right and explicit connections between unseen and seen relations, thereby building a better semantic space for ZSRC task.} 

\section{Conclusion and Future Work}

We have studied the zero-shot relation classification task and took the first step towards bridging symbolic reasoning with semantic representations. Extensive experiments demonstrate the efficacy of our approach, revealing the advantages of knowledge graph embeddings and rules. %We empirically observe that implicit and explicit semantic connections perform better than previous word embedding based methods, which may shed light on future work on zero-shot approaches. 
%We anticipate further research on promising directions for zero-shot relation classification work, including
In the future, we plan to  exploit more efficient approaches to obtain symbolic rules an build end-to-end reasoning approaches for zero-shot tasks.%; 3) investigating the model interpretation of zero-shot relation classification. 

\section*{Acknowledgments}
We  want to express gratitude to the anonymous reviewers for their hard work and kind comments, which will further improve our work in the future. This work is funded by NSFC91846204/U19B2027/61473260, national key research program 2018YFB1402800/SQ2018YFC000004, Alibaba CangJingGe (Knowledge Engine) Research Plan.

\bibliographystyle{coling}
\bibliography{coling2020}

\end{document}